# CONCEPTION ISOTROPIQUE D'UNE MORPHOLOGIE PARALLELE : APPLICATION A L'USINAGE

**D. CHABLAT, P. WENGER**
Institut de Recherche en Cybernétique de Nantes,
1 rue de la Noë, 44321 Nantes, France
**J. ANGELES**
Centre for Intelligent Machines, McGill University, 817 Sherbrooke Street West,
Montréal, Québec, Canada H3A 2K6

**Abstract** : The aim of this paper is the isotropic design of a hybrid morphology dedicated to 3-axis machining applications. It is necessary to ensure the feasibility of continuous, singularity-free trajectories, as well as a good manipulability in position and velocity. We want to propose an alternative design to conventional serial machine-tools. We compare a serial *PPP* machine-tool (three prismatic orthogonal axes) with a hybrid architecture which we optimize only the first two axes. The critrerion used for the optimization is the conditioning of the Jacobian matrices. The optimum, namely isotropy, can be obtained which provides our architecture with excellent manipulability properties.

**Résumé :** : Le but de cet article est la conception isotropique d'une morphologie hybride dont la vocation est l'usinage en trois axes. Pour cela, nous devons garantir la réalisation de trajectoires continues, c'est-à-dire sans singularité, et la manipulabilité en position et en vitesse. Pour présenter une alternative possible aux morphologies sérielles des machines outils traditionnelles, nous comparons une morphologie sérielle de type *PPP*, c'est-à-dire trois articulations prismatiques placées orthogonalement deux à deux, et une structure hybride dont nous allons optimiser les deux premiers axes de déplacement, c'est-à-dire la table de la machine outil. Nous utilisons, comme critère d'optimisation, le conditionnement des matrices jacobiennes. L'optimum obtenu, c'est-à-dire l'isotropie, donne à la structure une excellente manipulabilité en position et en vitesse.
.

## 1. Préliminaires

### 1.1. Morphologie sérielle à trois degrés de liberté

Dans l'industrie, les machines outils trois axes possèdent une morphologie sérielle simple. Pour réaliser les déplacements dans les trois directions de l'espace, on utilise trois actionneurs prismatiques, soit *PPP*, placés orthogonalement (Figure 1). Dans ce cas, il y a découplage entre le déplacement dans le plan *XY* et le déplacement suivant l'axe *Z* où se situe l'axe de rotation de l'outil.

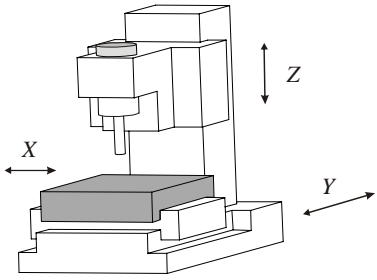

Figure 1 : Morphologie de machine outil 3 axes classique

Le problème de cette morphologie est que l'actionneur commandant l'axe des *X* supporte à la fois la pièce à usiner et l'actionneur commandant le déplacement de l'axe des *Y*. Ceci entraîne une asymétrie dans le comportement de la machine qui diminue ses performances cinématiques et surtout dynamiques. Pour résoudre ce problème, il est possible de changer la morphologie des machines outils en utilisant soit des morphologies parallèles soit des morphologies hybrides. Pour la morphologie *PPP*, le modèle cinématique est le suivant,

$$\mathbf{J}\,\dot{\boldsymbol{\rho}} = \dot{\mathbf{p}} \text{ avec } \mathbf{J} = \mathbf{1}_{3\times 3}$$

où $\dot{\mathbf{p}} = [\dot{x}\ \dot{y}\ \dot{z}]^T$ le vecteur vitesse de déplacement d'un point de l'outil *P* et $\dot{\boldsymbol{\rho}} = [\dot{\rho}_1\ \dot{\rho}_2\ \dot{\rho}_3]^T$ le vecteur vitesse des articulations prismatiques. La matrice jacobienne cinématique **J** étant unitaire, l'ellipsoïde de manipulabilité en vitesse et en force [Yoshikawa 85] est une sphère unitaire et cela pour toutes les configurations de l'espace de travail. L'objectif de cet article est de se rapprocher de ce modèle cinématique en utilisant une morphologie parallèle ou hybride.

## 1.2. Morphologie parallèle plane à deux degrés de liberté

Nous focalisons notre étude sur le remplacement de la table de la machine outil de type *PP*, par une structure parallèle. Ainsi, nous avons choisi d'étudier un manipulateur 5 barres (Figure 2).

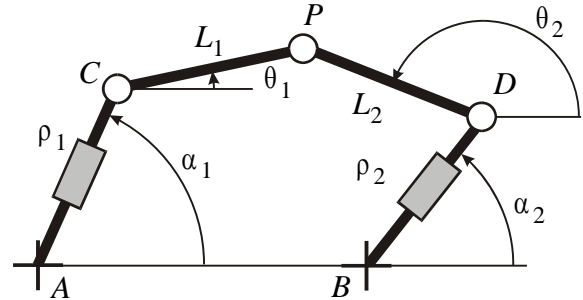

Figure 2 : Morphologie parallèle à deux degrés de liberté

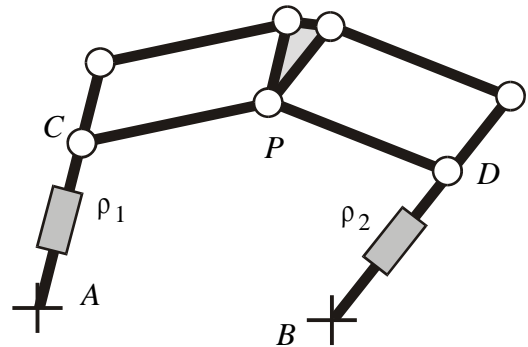

Figure 3 : Morphologie parallèle à deux degrés de liberté avec contrôle de l'orientation

Les variables articulaires sont les variables $\rho_1$ et $\rho_2$ associées aux deux actionneurs prismatiques et les variables de l'espace opérationnel sont la position du point $P = [x\ y]^T$. Les longueurs $L_1$ et $L_2$, les angles $\alpha_1$ et $\alpha_2$, et la position des points *A* et *B* définissent complètement la géométrie du manipulateur. Pour réduire le nombre de variables de conception, nous posons $L_1 = L_2$. Cette simplification permet de rendre symétrique la morphologie, de réduire le nombre de pièces différentes et donc de réduire son coût de fabrication.

Pour maintenir l'orientation du repère attaché au point *P*, nous choisissons d'utiliser deux parallélogrammes. Ainsi, nous augmentons la rigidité de la structure (Figure 3). Le dimensionnement des barres entrant dans la composition des parallélogrammes n'est pas pris en compte dans notre étude. Cependant, la rigidité de la structure ainsi que les limites articulaires des articulations passives sont fonction des solutions technologiques retenues pour leur conception.

Pour obtenir le troisième axe de la machine outil, il est possible de placer une troisième articulation prismatique orthogonalement aux deux premières. Celle-ci peut être située découplée comme dans le cas de la Figure 1.

### 1.3. Cinématique de la morphologie parallèle plane

La vitesse $\dot{\mathbf{p}}$ du point *P* peut être écrite de deux manières différentes. Ainsi, en parcourant la boucle fermée (*ACP-BDP*) dans les deux sens possibles, nous obtenons

$$\dot{\mathbf{p}} = \dot{\mathbf{c}} + \dot{\theta}_1 \mathbf{E} (\mathbf{p} - \mathbf{c}) \qquad (1a)$$

$$\dot{\mathbf{p}} = \dot{\mathbf{d}} + \dot{\theta}_2 \mathbf{E} (\mathbf{p} - \mathbf{d}) \qquad (1b)$$

où

- **E** est la matrice de rotation à 90° définie dans le plan,

$$\mathbf{E} = \begin{bmatrix} 0 & -1 \\ 1 & 0 \end{bmatrix};$$

- **c** et **d** représentent le vecteur position des points *C* et *D*, respectivement.

De plus, les vitesses $\dot{\mathbf{c}}$ et $\dot{\mathbf{d}}$ des points *C* et *D* sont données par,

$$\dot{\mathbf{c}} = \frac{\mathbf{c} - \mathbf{a}}{\|\mathbf{c} - \mathbf{a}\|} \dot{\rho}_1 = \begin{bmatrix} \cos(\alpha_1) \\ \sin(\alpha_1) \end{bmatrix} \dot{\rho}_1,$$

$$\dot{\mathbf{d}} = \frac{\mathbf{d} - \mathbf{b}}{\|\mathbf{d} - \mathbf{b}\|} \dot{\rho}_2 = \begin{bmatrix} \cos(\alpha_2) \\ \sin(\alpha_2) \end{bmatrix} \dot{\rho}_2$$

où les angles $\alpha_1$ et $\alpha_2$ sont les orientations des actionneurs prismatiques par rapport à l'horizontale.

Maintenant, nous allons éliminer les vitesses articulaires $\dot{\theta}_1$ et $\dot{\theta}_2$ des équations (1a) et (1b) en les multipliant par $(\mathbf{p} - \mathbf{c})^T$ et $(\mathbf{p} - \mathbf{d})^T$.

$$(\mathbf{p} - \mathbf{c})^T \dot{\mathbf{p}} = (\mathbf{p} - \mathbf{c})^T \frac{\mathbf{c} - \mathbf{a}}{\|\mathbf{c} - \mathbf{a}\|} \dot{\rho}_1 \qquad (2a)$$

$$(\mathbf{p} - \mathbf{d})^T \dot{\mathbf{p}} = (\mathbf{p} - \mathbf{d})^T \frac{\mathbf{d} - \mathbf{b}}{\|\mathbf{d} - \mathbf{b}\|} \dot{\rho}_2 \qquad (2b)$$

Nous regroupons les équations (2a) et (2b) sous une forme vectorielle,

$$\mathbf{A} \dot{\mathbf{p}} = \mathbf{B} \dot{\boldsymbol{\rho}}$$

avec **A** et **B** indiquant les matrices jacobiennes parallèle et sérielle respectivement. On note :

$$\mathbf{A} \equiv \begin{bmatrix} (\mathbf{p} - \mathbf{c})^T \\ (\mathbf{p} - \mathbf{d})^T \end{bmatrix},$$

$$\mathbf{B} \equiv \begin{bmatrix} (\mathbf{p} - \mathbf{c})^T \frac{\mathbf{c} - \mathbf{a}}{\|\mathbf{c} - \mathbf{a}\|} & 0 \\ 0 & (\mathbf{p} - \mathbf{d})^T \frac{\mathbf{d} - \mathbf{b}}{\|\mathbf{d} - \mathbf{b}\|} \end{bmatrix},$$

$$\dot{\boldsymbol{\rho}} = \begin{bmatrix} \dot{\rho}_1, \dot{\rho}_2 \end{bmatrix} \text{ et } \dot{\mathbf{p}} = \begin{bmatrix} \dot{x}, \dot{y} \end{bmatrix}$$

En étudiant les matrices **A** et **B**, nous pouvons définir les singularités parallèles et sérielles de ce manipulateur. Lorsque **A** et **B** sont inversibles, nous pouvons de même étudier la matrice jacobienne cinématique **J** [Merlet 97] pour optimiser le manipulateur,

$$\dot{\mathbf{p}} = \mathbf{J} \dot{\boldsymbol{\rho}} \quad \text{avec} \quad \mathbf{J} = \mathbf{A}^{-1} \mathbf{B} \qquad (3a)$$

où la matrice jacobienne cinématique inverse $\mathbf{J}^{-1}$, telle que

$$\dot{\boldsymbol{\rho}} = \mathbf{J}^{-1} \dot{\mathbf{p}} \quad \text{avec} \quad \mathbf{J}^{-1} = \mathbf{B}^{-1} \mathbf{A} \qquad (3b)$$

## 1.4. Singularités parallèles

Les *singularités parallèles* sont dues à la perte de rang de la matrice jacobienne parallèle **A** [Chablat 98], c'est-à-dire lorsque det(**A**) = 0. Dans ce cas, il est possible de déplacer localement la plate-forme mobile alors que les articulations motorisées sont bloquées. Ces singularités sont particulièrement néfastes, car les efforts au sein de la structure s'accroissent dangereusement. Pour éviter toute détérioration, il est nécessaire de limiter les variations articulaires.

Pour la morphologie étudiée, les singularités parallèles apparaissent lorsque les points *C*, *D*, et *P* sont alignés (Figure 4).

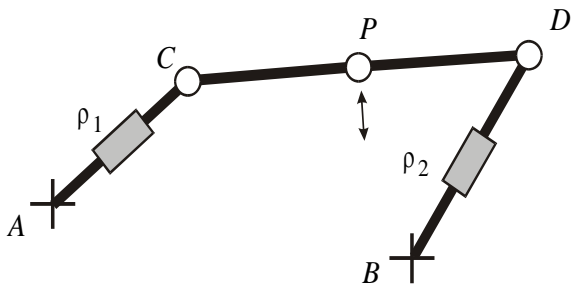

Figure 4 : Singularité parallèle

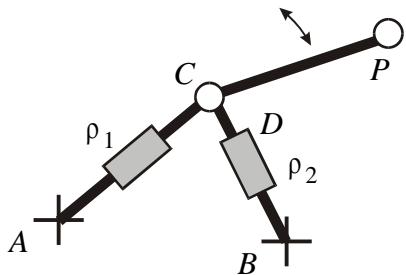

Figure 5 : Singularité structurelle

Ces postures se produisent lorsque $\theta_1 - \theta_2 = k\pi$, pour *k* entier. Elles se situent à l'intérieur de l'espace de travail et forment les limites de l'ensemble articulaire. De plus, des singularités structurelles peuvent se produire lorsque $L_1$ est égale à $L_2$ (Figure 5). Dans ces configurations, la commande de l'effecteur est perdue.

## 1.5. Singularités sérielles

Les *singularités sérielles* sont dues à la perte de rang de matrice jacobienne sérielle **B**, c'est-à-dire lorsque det(**B**) = 0. Dans ce cas, il n'est pas possible de réaliser certaines vitesses de la plate-forme mobile. Les singularités sérielles représentent les limites de l'espace de travail **[Merlet 97]**.

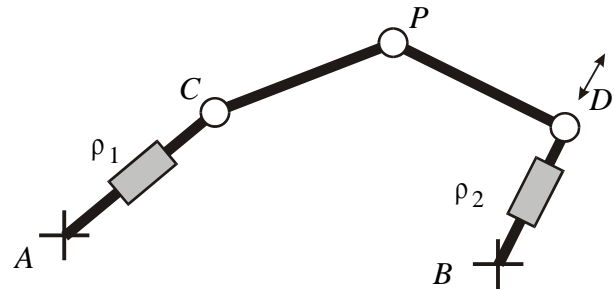

Figure 6 : Singularité sérielle

Pour la morphologie étudiée, les singularités sérielles apparaissent lorsque $\theta_1 - \alpha_1 = \pi/2 + k\pi$, , pour *k* entier ou lorsque $\theta_2 - \alpha_2 = \pi/2 + k\pi$, pour *k* entier (Figure 6). Elles traduisent une perte de manipulabilité en force.

## 1.6. Application à l'usinage

Pour une machine outil trois axes, le déplacement de la table se fait suivant deux axes perpendiculaires. La course de chaque actionneur donne la dimension de l'espace de travail. Pour les manipulateurs parallèles, cette transformation n'est pas directe. Il en résulte souvent un espace de travail dont les dimensions sont beaucoup plus petites. L'objectif de notre conception est de disposer d'un espace de travail équivalent à celui d'une machine outil classique. Dans ce cas, nous souhaitons disposer d'un espace de travail dont la forme se rapproche de la forme d'une table de machine outil c'est-à-dire possédant une forme rectangulaire.

## 2. Conception isotropique

L'objectif de notre conception est de supprimer les singularités parallèles de l'espace articulaire. Ainsi, la commande de la machine outil sera simplifiée. Cependant, nous souhaitons conserver de bonnes propriétés cinématiques. Nous allons dans un premier temps, isoler les conditions d'isotropie des matrices jacobiennes puis faire une étude comparative entre l'espace de travail et l'ensemble articulaire d'une morphologie classique de type Biglide et la morphologie ainsi définie. La conception isotropique de manipulateurs parallèles a déjà été étudié pour les manipulateurs parallèles plans [Daniali 95] et sphériques [Gosselin 88].

### 2.1. Conditionnement d'une matrice homogène

Au cours de notre processus de conception, nous allons définir les courbes d'iso-conditionnement des matrices jacobiennes. Nous étudions le conditionnement $\kappa(\mathbf{M})$ d'une matrice $\mathbf{M}$ de dimension $m \times n$, avec $m \leq n$, comme étant le rapport entre la plus grande valeur singulière $\sigma_g$ et la plus petite valeur singulière $\sigma_p$ de la matrice $\mathbf{M}$ [Golub 89] :

$$\kappa(\mathbf{M}) = \frac{\sigma_g}{\sigma_p}$$

Les valeurs singulières $\{\sigma_k\}_{1,m}$ de la matrice $\mathbf{M}$ sont définies comme étant les racines carrées non-négatives des valeurs propres de la matrice semi-définie positive $\mathbf{M}\mathbf{M}^T$ de dimension $m \times m$.

### 2.2. Conditionnement de la matrice jacobienne parallèle

Pour calculer le conditionnement de la matrice jacobienne parallèle, il est nécessaire de calculer $\mathbf{A}\mathbf{A}^T$ :

$$\mathbf{A}\mathbf{A}^T = L_1^2 \begin{bmatrix} 1 & \cos(\theta_1 - \theta_2) \\ \cos(\theta_1 - \theta_2) & 1 \end{bmatrix}$$

Les valeurs propres $\eta_1$ et $\eta_2$ de $\mathbf{A}\mathbf{A}^T$ sont alors :

$$\eta_1 = L_1^2 (1 + \cos(\theta_1 - \theta_2)) \text{ et}$$
$$\eta_2 = L_1^2 (1 - \cos(\theta_1 - \theta_2))$$

Nous définissons le conditionnement de la matrice jacobienne parallèle $\mathbf{A}$ comme :

$$\kappa(\mathbf{A}) = \sqrt{\frac{\eta_{max}}{\eta_{min}}}$$

où $\eta_{min} = 1 - |\cos(\theta_1 - \theta_2)|$ et $\eta_{max} = 1 + |\cos(\theta_1 - \theta_2)|$. Après simplification, nous obtenons l'expression suivante :

$$\kappa(\mathbf{A}) = \frac{1}{|\tan((\theta_2 - \theta_1)/2)|}$$

Ainsi, le conditionnement de la matrice jacobienne parallèle est minimum, c'est-à-dire égal à 1, lorsque $|\theta_1 - \theta_2| = \pi/2 + k\pi$, pour $k$ entier. Inversement, le conditionnement de la matrice jacobienne parallèle tend vers l'infini lorsque $|\theta_1 - \theta_2| = k\pi$, pour $k$ entier. Les configurations pour lesquelles $\kappa(\mathbf{A}) = 1$ sont appelées configurations isotropes (Figure 7), alors que les configurations pour lesquelles $\kappa(\mathbf{A}) \to +\infty$ sont les singularités parallèles du manipulateur (Figure 4).

L'étude du conditionnement de la matrice $\mathbf{A}$ nous permet de trouver un ensemble de configurations du manipulateur pour lesquelles la matrice $\mathbf{A}$ est isotrope. Ces conditions ne portent pas sur les orientations $\alpha_1$ et $\alpha_2$ des actionneurs.

## 2.3. Conditionnement de la matrice jacobienne sérielle

La matrice **B** étant diagonale, ses valeurs singulières, $\beta_1$ et $\beta_2$, sont les valeurs absolues des valeurs de la diagonale. Le conditionnement $\kappa(\mathbf{B})$ de la matrice **B** est défini tel que :

$$\kappa(\mathbf{B}) = \sqrt{\frac{\beta_{max}}{\beta_{min}}}$$

$$\text{avec } \mathbf{B} \equiv L_1 \begin{bmatrix} \cos(\theta_1 - \alpha_1) & 0 \\ 0 & \cos(\theta_2 - \alpha_2) \end{bmatrix}$$

où $\beta_{max}$ et $\beta_{min}$ sont définis comme suit : si $|\cos(\theta_1 - \alpha_1)| < |\cos(\theta_2 - \alpha_2)|$ alors $\beta_{min} = |\cos(\theta_1 - \alpha_1)|$ et $\beta_{max} = |\cos(\theta_2 - \alpha_2)|$ ; sinon, $\beta_{min} = |\cos(\theta_2 - \alpha_2)|$ et $\beta_{max} = |\cos(\theta_1 - \alpha_1)|$

Le conditionnement de la matrice jacobienne sérielle est minimum, soit $\kappa(\mathbf{B}) = 1$, lorsque $|\cos(\theta_1 - \alpha_1)| = |\cos(\theta_2 - \alpha_2)| \neq 0$.

Inversement, il tend vers l'infini lorsque $|\cos(\theta_1 - \alpha_1)| = 0$ ou lorsque $|\cos(\theta_2 - \alpha_2)| = 0$. Les configurations pour lesquelles $\kappa(\mathbf{B}) = 1$ sont les configurations isotropes (Figure 8) et les configurations pour lesquelles $\kappa(\mathbf{B}) \to +\infty$ sont les singularités sérielles du manipulateur (Figure 6). Comme pour l'étude du conditionnement de la matrice **A**, les conditions d'isotropie de la matrice **B** ne portent pas sur l'orientation $\alpha_1$ et $\alpha_2$ des actionneurs.

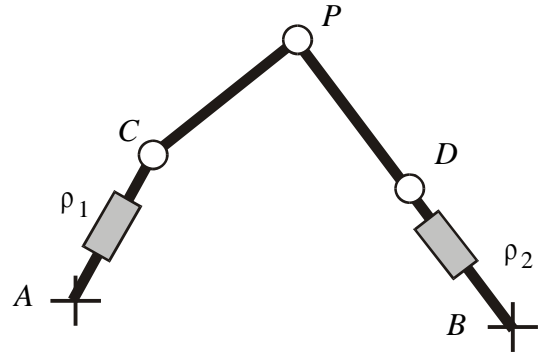

Figure 7 : Configuration isotrope de la matrice jacobienne parallèle

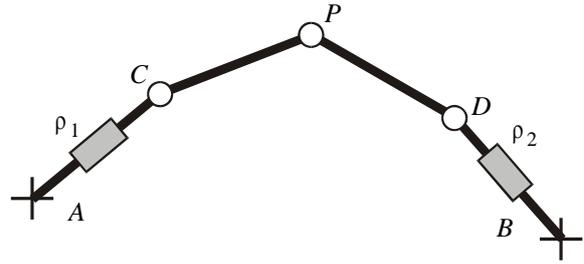

Figure 8 : Configuration isotrope de la matrice jacobienne sérielle

## 2.4. Conditionnement de la matrice jacobienne cinématique

Pour définir l'orientation des articulations prismatiques, nous allons étudier le conditionnement de la matrice jacobienne cinématique inverse $\mathbf{J}^{-1}$ donnée dans l'équation (3b). Dans ce cas, les matrices $\mathbf{B}^{-1}$ et $\mathbf{J}^{-1}$ s'écrivent simplement,

$$\mathbf{B}^{-1} = \frac{1}{L_1} \begin{bmatrix} \frac{1}{c_1} & 0 \\ 0 & \frac{1}{c_2} \end{bmatrix}, \mathbf{J}^{-1} = \begin{bmatrix} \frac{1}{c_1}(\mathbf{p} - \mathbf{c})^T \\ \frac{1}{c_2}(\mathbf{p} - \mathbf{d})^T \end{bmatrix}$$

$$\text{avec } c_i = \cos(\theta_i - \alpha_i), i = 1, 2$$

Les conditions d'isotropie de la matrice $\mathbf{J}^{-1}$ sont les suivantes,

$$\text{i) } \frac{1}{c_1} \|\mathbf{p} - \mathbf{c}\| = \frac{1}{c_2} \|\mathbf{p} - \mathbf{d}\| \quad \text{et}$$

$$\text{ii) } (\mathbf{p} - \mathbf{c})^T (\mathbf{p} - \mathbf{d}) = 0$$

Dans la suite de cet article, nous posons $\alpha_1 = 0$ et $\alpha_2 = \pi / 2$ afin de rendre $\mathbf{J}^{-1}$ unitaire sur la configuration isotrope lorsque $\theta_1 = 0$ et $\theta_2 = \pi / 2$. Dans cette configuration, un déplacement de l'articulation prismatique lié à $\rho_1$ produit un mouvement colinéaire à *x* (Figure 9) et un déplacement de l'articulation prismatique lié à $\rho_2$ produit un mouvement colinéaire à *y* comme c'est le cas sur une machine outil d'architecture *PP*. De plus, si nous plaçons les coordonnées du point *A* en (0 0) puis *B* en (M -M) (Figure 10), les variables articulaires $\rho_1$ et $\rho_2$ sont égales lorsque le manipulateur atteint sa configuration isotrope.

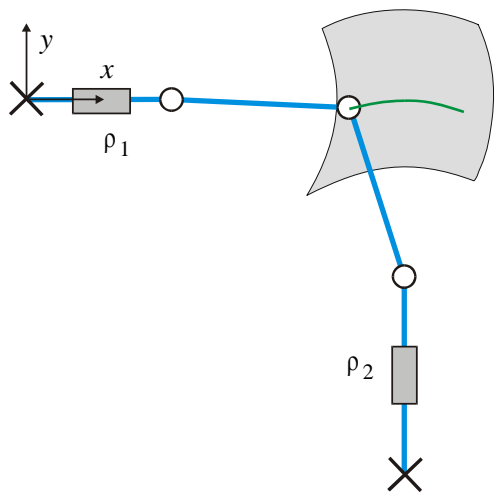

Figure 9 : Déplacement du point *P* engendré par le déplacement d'une articulation prismatique

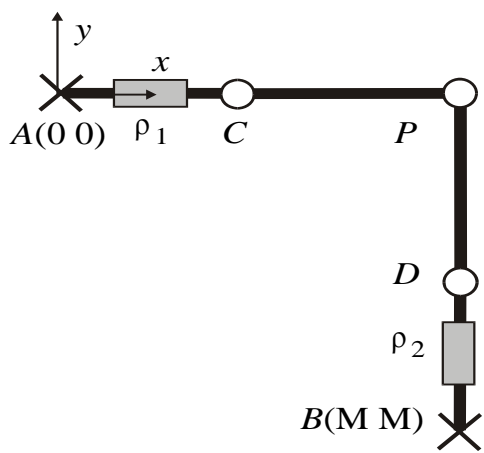

Figure 10 : Configuration isotrope

Cette constatation nous permet d'aborder le problème d'amplification de vitesse dû à l'utilisation d'une morphologie parallèle. En effet, dans le cas d'une morphologie sérielle possédant trois axes de déplacement linéaire, le déplacement d'une articulation provoque le même déplacement de l'outil (ou de la pièce). Pour les morphologies parallèles, ces déplacements ne sont pas équivalents. Lorsque le manipulateur est proche d'une singularité parallèle, il se produit une multiplication des déplacements, c'est-à-dire que le déplacement d'une unité de mesure d'une articulation peut provoquer un déplacement plusieurs fois supérieures dans l'espace de travail. Inversement, pour usiner à une précision constante, la précision des actionneurs doit être très grande.

## 2.5. Étude de l'espace de travail et de l'ensemble articulaire

Dans cette étude, nous allons voir que les conditions d'isotropie donnent au manipulateur étudié des propriétés très intéressantes. Pour cela, nous allons étudier deux morphologies, l'une où les orientations des actionneurs sont $\alpha_1 = \alpha_2 = 0$ de type biglide (Figure 11) et l'autre avec les orientations précédemment calculées (Figure 12).

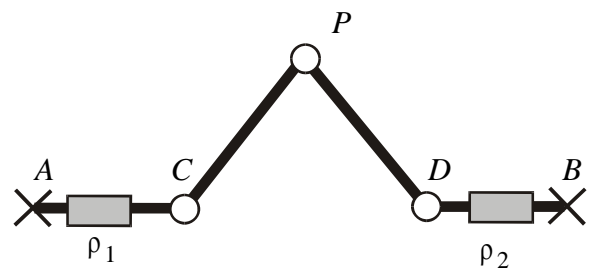

Figure 11 : Morphologie de type biglide

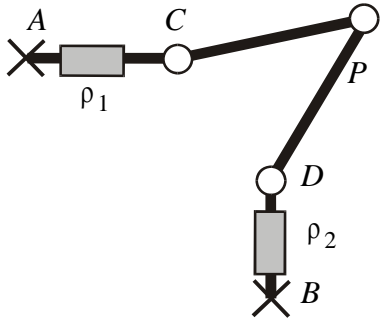

Figure 12 : Morphologie isotrope

Pour définir les limites de l'ensemble articulaire, nous allons étudier les ellipsoïdes de manipulabilité en vitesse issues de la matrice $\mathbf{J}^{-1}$ [Yoshikawa 85]. En utilisant l'équation (3b), nous pouvons écrire une relation entre la vitesse $\dot{\mathbf{p}}$ du point $P$ et la vitesse articulaire $\dot{\boldsymbol{\rho}}$. En posant $\|\dot{\boldsymbol{\rho}}\| \leq 1$, nous obtenons :

$$\dot{\mathbf{p}}^T (\mathbf{J}\mathbf{J}^T)^{-1} \dot{\mathbf{p}} \leq 1 \qquad (4)$$

L'équation (4) définit le domaine de variation de $\dot{\mathbf{p}}$. La transformation d'un cercle de rayon unitaire de l'ensemble articulaire par la matrice $(\mathbf{J}\mathbf{J}^T)^{-1}$ donne une ellipse dans l'espace de travail (Figure 13) [Lallemand 94].

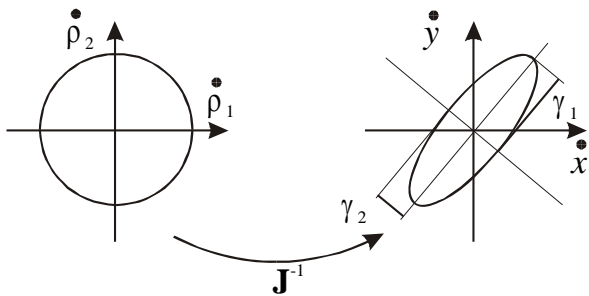

Figure 13 : Définition de la manipulabilité en vitesse

Les racines carrées $\gamma_1$ et $\gamma_2$ des valeurs propres de la matrice $(\mathbf{J}\mathbf{J}^T)^{-1}$ sont les valeurs des demi-axes de l'ellipse qui définissent les deux facteurs d'amplification de vitesse (actionneur vers effecteur), $\lambda_1 = 1/\gamma_1$ et $\lambda_2 = 1/\gamma_2$, selon ces axes principaux. Pour limiter les variations de ces facteurs dans l'espace de travail, nous posons comme contrainte que,

$$1/3 < \lambda_i < 3 \qquad (5)$$

Cela signifie que lorsque l'on impose une vitesse $\mathbf{v}$ dans l'espace articulaire, la vitesse résultante dans l'espace de travail est au plus trois fois plus grande ou au moins trois fois plus petite. Ce critère est très important car il conditionne aussi la précision en position de l'outil de la machine.

Pour pouvoir comparer les deux manipulateurs, la distance entre les deux points $A$ et $B$ est constante ainsi que la longueur des barres $L_1$ et $L_2$. De même, pour améliorer la comparaison, l'échelle de représentation de l'ensemble articulaire et de l'espace de travail est identique.

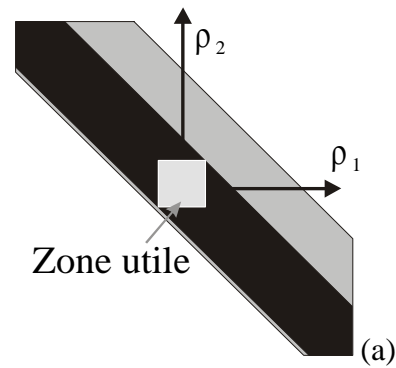

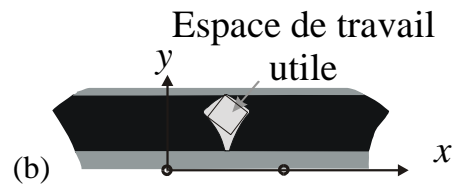

Figure 14 : Ensemble articulaire (a) et espace de travail (b) pour la morphologie biglide

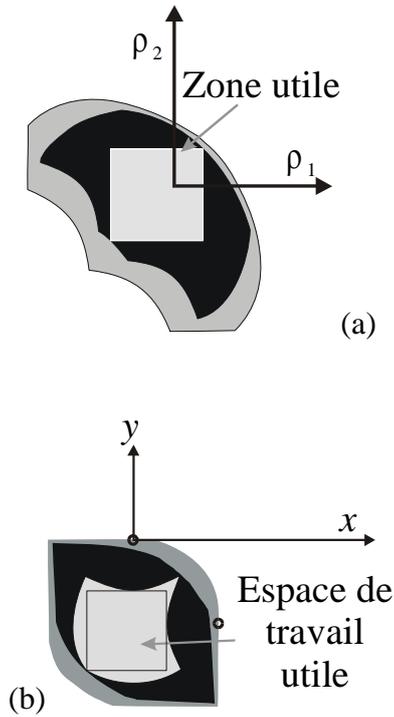

Figure 15 : Ensemble articulaire (a) et espace de travail (b) pour la morphologie isotrope

L'objectif premier de notre recherche était de concevoir un manipulateur ne possédant pas de singularité dans son espace de travail. Pour cela, nous allons utiliser les contraintes imposées par l'équation (5) pour définir la *zone utile* de forme carré dans l'ensemble articulaire et son image dans l'espace de travail, *l'espace de travail utile*.

Dans le cas de la morphologie isotrope, la définition d'une zone sans singularité dans l'ensemble articulaire et respectant les contraintes sur les facteurs d'amplification de vitesse nous conduit à placer des butées articulaires moins sévères que dans le cas de la morphologie Biglide (Figures 14a et 15a). De plus, l'espace de travail utile de la morphologie isotrope est plus adapté à l'usinage car le carré inscrit est 7 fois plus grand que celui de la morphologie biglide (Figures 14b et 15b).

Pour la morphologie isotrope, nous montrons dans la Figure 16 les courbes d'iso-valeurs des facteurs d'amplification de vitesse.

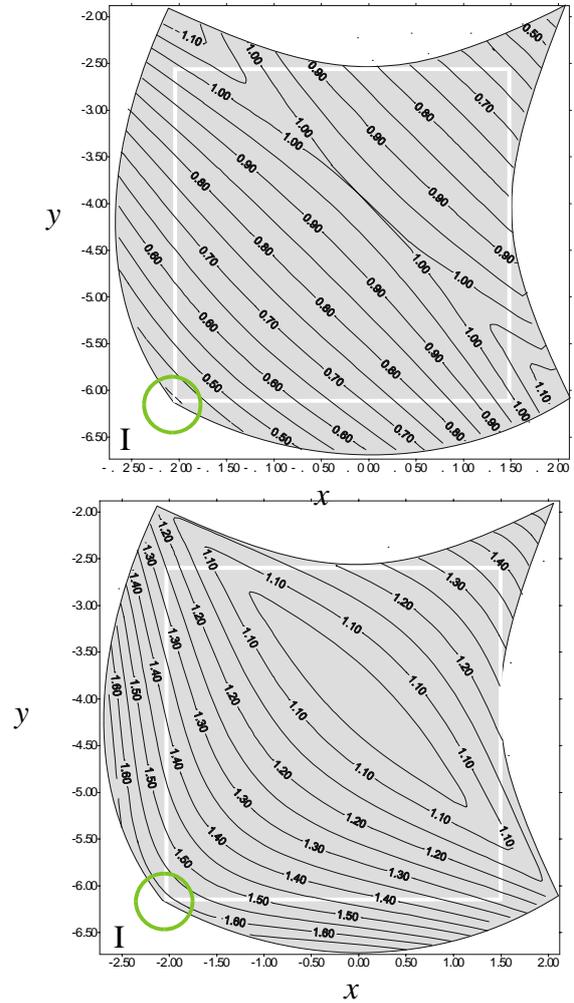

Figure 16 : Courbes d'iso-valeurs des facteurs d'amplification de vitesse $\lambda_i$ dans l'espace de travail utile de la morphologie isotrope

On constate alors que les contraintes d'amplification de vitesse $\lambda_i$ varient entre [0.4 , 1.6]. De plus, si l'on délimite une zone carrée dans l'espace de travail utile, on observe que les contraintes ne sont saturées que dans une petite zone (noté I) placée sur la frontière.

## 3. Conclusion

Nous venons de réaliser une conception isotropique d'une morphologie parallèle à deux degrés de liberté. Nous constatons que l'optimisation des dimensions en étudiant le conditionnement des matrices jacobiennes permet d'obtenir à la fois de bonnes performances cinématiques et un bon espace de travail. En considérant l'ensemble articulaire, il est possible de simplifier la commande en supprimant les singularités parallèles. De plus, l'utilisation du facteur d'amplification de vitesse permet de définir une distance par rapport aux singularités parallèles.

Par la suite, nous souhaitons généraliser cette morphologie au cas spatial, pour obtenir une morphologie parallèle à trois degrés de liberté en position.

## Bibliographie